\documentclass[runningheads,orivec]{llncs}
\usepackage[T1]{fontenc}
\usepackage{graphicx}
\usepackage{hyperref}
\usepackage{framed,multirow}

\usepackage{amssymb}
\usepackage{latexsym}

\usepackage{url}
\usepackage{xcolor}

\usepackage{hyperref}

\usepackage{amsmath,amssymb,amsfonts}
\usepackage{algorithmic}
\usepackage{graphicx}
\usepackage{textcomp}
\usepackage{subfigure}
\usepackage{algorithm}
\usepackage{graphics}
\usepackage{threeparttable}
\usepackage{color}
\usepackage[normalem]{ulem}
\usepackage{multirow}
\usepackage{float}
\usepackage{amsfonts}
\usepackage{bm}
\usepackage{array}
\usepackage{colortbl}
\usepackage{pifont}
\usepackage{diagbox}
\usepackage{rotating}
\usepackage{booktabs}
\usepackage{overpic}
\usepackage{makecell}
\usepackage{contour}
\usepackage{mathtools}
\usepackage{gensymb}
\usepackage{marvosym}
\usepackage[square,sort,comma,numbers]{natbib}

\definecolor{newcolor}{rgb}{.8,.349,.1}

\begin{document}
\title{Explainable fetal ultrasound quality assessment with progressive concept bottleneck models}
\titlerunning{Progressive concept bottleneck models}
% \author{Anonymous authors}
% \institute{Anonymous institute}
\author{Manxi 
Lin\inst{1},
Aasa Feragen\inst{1}\textsuperscript{(\Letter)},
Kamil Mikolaj\inst{1},
Zahra Bashir\inst{2,3},
\\ Morten B. S. Svendsen\inst{1,3},
Martin G. Tolsgaard\inst{3,4,5},
\\
Anders N. Christensen\inst{1}
}
\institute{
$^1$~Technical University of Denmark, Kongens Lyngby, Denmark\\
\email{afhar@dtu.dk}\\
$^2$~Slagelse Hospital, Slagelse, Denmark\\
$^3$~CAMES, Copenhagen, Denmark\\
$^4$ Copenhagen University Hospital Rigshospitalet, Copenhagen, Denmark\\
$^5$ Department of Clinical Medicine, University of Copenhagen, Copenhagen, Denmark
}
\authorrunning{Lin et al.}
\maketitle              % typeset the header of the contribution
\begin{abstract}
The quality of fetal ultrasound screening scans directly influences the precision of biometric measurements. However, acquiring high-quality scans is labor-intensive and highly relies on the operator's skills. Considering the low contrastiveness and imaging artifacts that widely exist in ultrasound, even a dedicated deep-learning model can be vulnerable to learning from confounding information in the image. In this paper, we propose a holistic and explainable method for fetal ultrasound quality assessment, where we design a hierarchical concept bottleneck model by introducing human-readable ``concepts" into the task and imitating the sequential expert decision-making process. This hierarchical information flow forces the model to learn concepts from semantically meaningful areas: The model first passes through a layer of visual, segmentation-based concepts, and next a second layer of property concepts directly associated with the decision-making task. We consider the quality assessment to be in a more challenging but more realistic setting, with fine-grained image recognition. Experiments show that our model outperforms equivalent concept-free models on an in-house dataset, and shows better generalizability on two public benchmarks, one from Spain and one from Africa, without any fine-tuning. Our implementation is publicly available~\href{https://github.com/mmmmimic/Progressive-Concept-Bottleneck-Models}{here}.

\end{abstract}
%
%
%

\section{Introduction}

\label{sec:introduction}
Ultrasound (US) imaging is widely used to screen fetal growth, due to its portability, low cost, and non-invasive nature~\citep{santos2017raman}. The evaluation of fetal US image quality is essential for accurately measuring biometric parameters in obstetric examinations~\citep{ambroise2023accuracy}. During screening, the clinician aims to obtain the best possible image quality, as quantified by the International Society of Ultrasound in Obstetrics and Gynecology (ISUOG). Their criteria for a high-quality scan, referred to as a \emph{standard plane} (SP), specify a set of requirements for anatomical structures that should, or should not, be visible in the image~\citep{salomon2019isuog}. Table~\ref{tab:isuog} summarizes the ISUOG criteria for third-trimester growth scans. Acquiring standard planes that satisfy these criteria is labor-intensive, and highly relies on the skill of the operators.

\begin{table*}[ht]
  \begin{center}
    \caption{ISUOG practical guidelines for the four 3rd-trimester standard anatomical planes: The cephalic (head); abdominal; femoral and cervix planes. The guidelines are summarized from~\citep{salomon2019isuog, kagan2015measure}.}
  \label{tab:isuog}
  \resizebox{\textwidth}{!}{
  \begin{tabular}{@{}llll@{}}
    \toprule
    Cephalic (head) plane & Abdominal plane & Femoral plane & Cervix plane\\
    \midrule
    symmetrical plane;
 & symmetrical plane; & \makecell[l]{both ends of bone clearly \\visible;} & \makecell[l]{cervix occupies $50-75\%$ \\of the total image;}\\
   plane showing thalamus; & plane showing stomach bubble; & $<45\degree$ angle to horizontal; & the bladder is empty; \\
   \makecell[l]{plane showing cavum\\ septi pellucidi;} & plane showing portal sinus; & \makecell[l]{femur occupying more than \\half of total image;} & \makecell[l]{the cervix is symmetric (no \\signs of excessive pressure);}\\
 cerebellum not visible; & kidneys not visible; & \makecell[l]{calipers placed correctly.}& \makecell[l]{the cervical canal is \\visualized sufficiently;}\\
 
 \makecell[l]{head occupying more \\than half of total image;}  & \makecell[l]{abdomen occupying more \\ than half of total image;}&  &\makecell[l]{calipers placed correctly at \\the internal and external \\orificium.}\\
 
  \makecell[l]{calipers and dotted \\ellipse placed correctly.}  & \makecell[l]{calipers and dotted \\ellipse placed correctly.}&  &\\
\bottomrule
  \end{tabular}
  }
  \end{center}
\end{table*}

\begin{figure}
\centering
\includegraphics[width=0.9\linewidth]{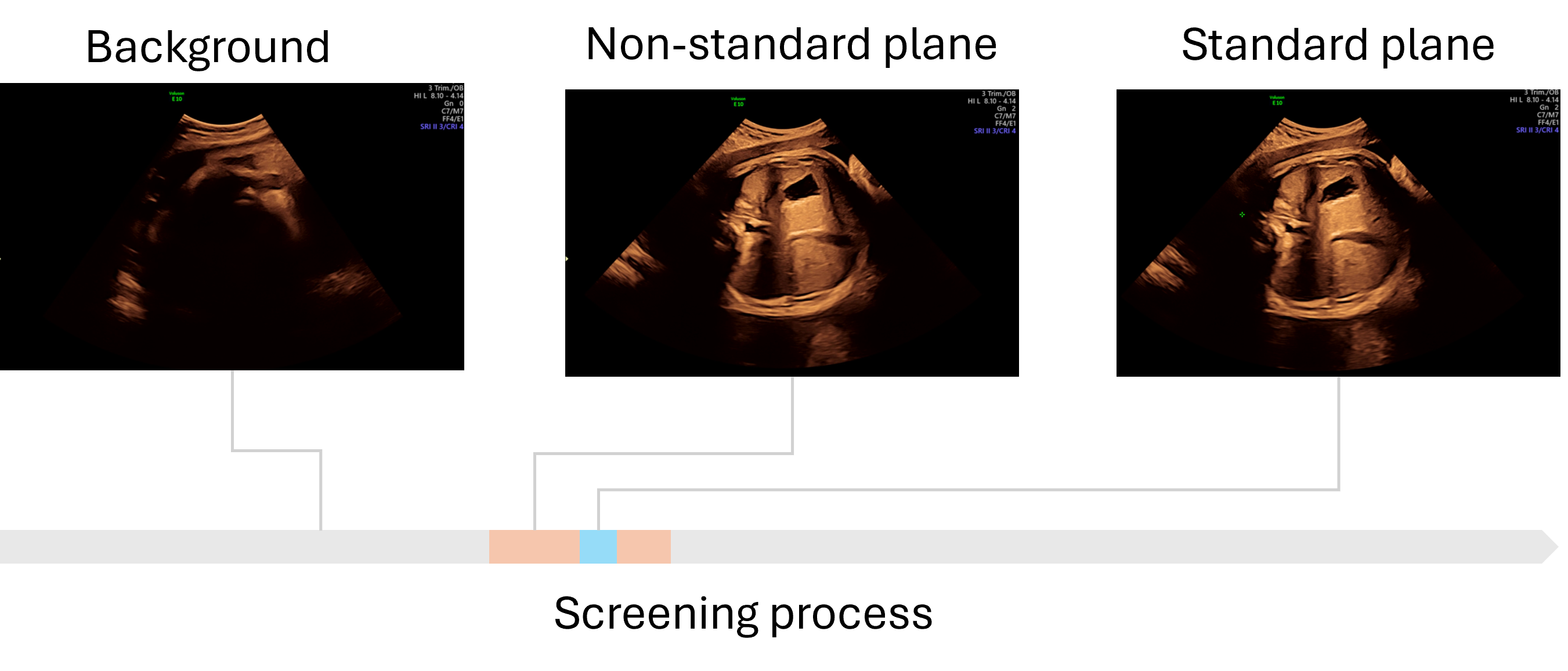}
\caption{Examples of an abdomen standard plane and two different definitions of the negative category.}
\label{fig:screen}
\end{figure}

Fig.~\ref{fig:screen} illustrates the image acquisition process during screening. The first observed ``background'' image does not show the required anatomical structures, whereas the next observed `non-standard plane' (NSP) refers to images showing the correct anatomy, but not fulfilling all ISUOG criteria; as such these two images represent different types of ''negative`` cases. During screening, the operator needs to fine-tune the pose of the probe to obtain the final SP, which is challenging considering its subtle difference from the NSP.

\begin{figure}[h!]
\centering
\includegraphics[width=\linewidth]{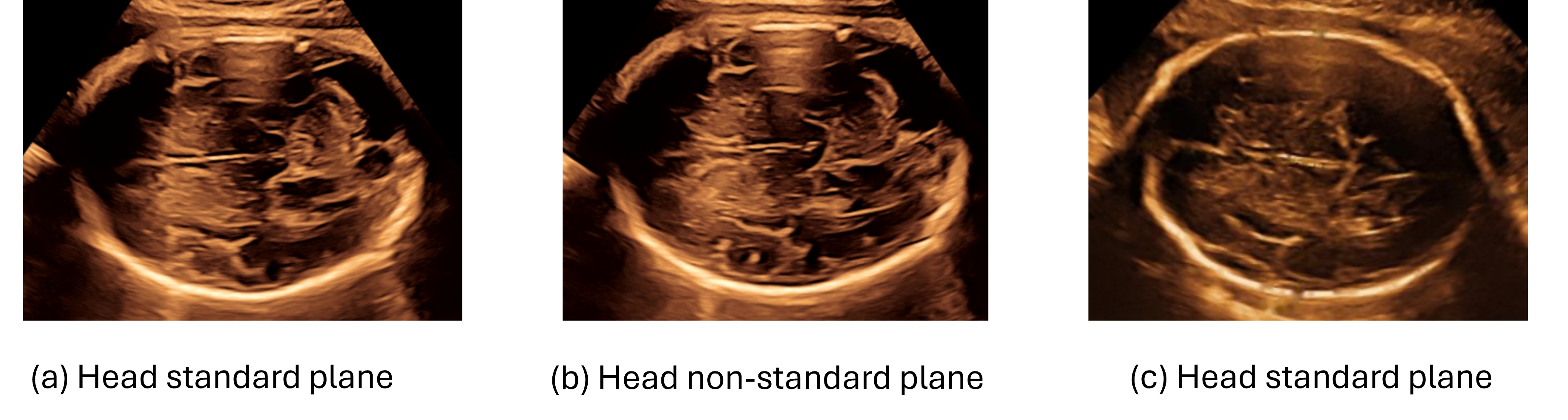}
\caption{Examples of head standard planes and non-standard planes.}
\label{fig:femurs}
\end{figure}

Deep learning is widely used for US image analysis~\citep{wu2017fuiqa,burgos2020evaluation,taksoee2024ai,guo2022fetal,dapueto2024knowledge}. 
However, many existing papers on ``standard plane recognition'' are designed to recognize US SPs from `background' images and SPs from different anatomical planes~\citep{dapueto2024knowledge,baumgartner2017sononet,droste2019ultrasound}. In this paper, we consider a more challenging but also more realistic task: We formulate standard plane recognition as a US quality assessment task, performing a fine-grained image recognition to distinguish SPs from NSPs across and in-between anatomical planes. This is challenging, as illustrated by the three examples shown in Fig.~\ref{fig:femurs}, due to the high intra-class variation between two SPs ((a) and (c)) compared to the low inter-class variation between SP and NSP ((a) and (b)).

The black-box nature of deep neural networks (DNNs) leads to potentially poorly grounded decisions, and therefore unacceptable risk. Increasing efforts are directed towards the explainability of model decisions. In contrast to post-hoc explanations~\citep{sundararajan2020many, wang2022hint, olah2017feature, kim2018interpretability, zhou2016cvpr}, concept bottleneck models (CBMs)~\citep{koh2020concept} aim to explain the AI decision using human specialist-predefined concepts, such as colors and shapes of objects, which are precise and accurate for human understanding. When considering a problem with input $x$ and target $y$, CBMs predict $y$ from expert-annotated concept labels. Specifically, given the concepts $c \in {\mathbb{R}}^{d}$, CBMs learn the mappings $x \xmapsto{g} c \xmapsto{f} y$ sequentially, jointly or independently~\citep{koh2020concept}. Here, $d$ is the dimension of the concept bottleneck. When the prediction of $y$ is based on human-interpretable concepts $c$, the model is inherently explainable.

\subsubsection{Holistic and explainable fetal US analysis.} 
\begin{figure}[t]
  \centering
  \includegraphics[width=0.9\linewidth]{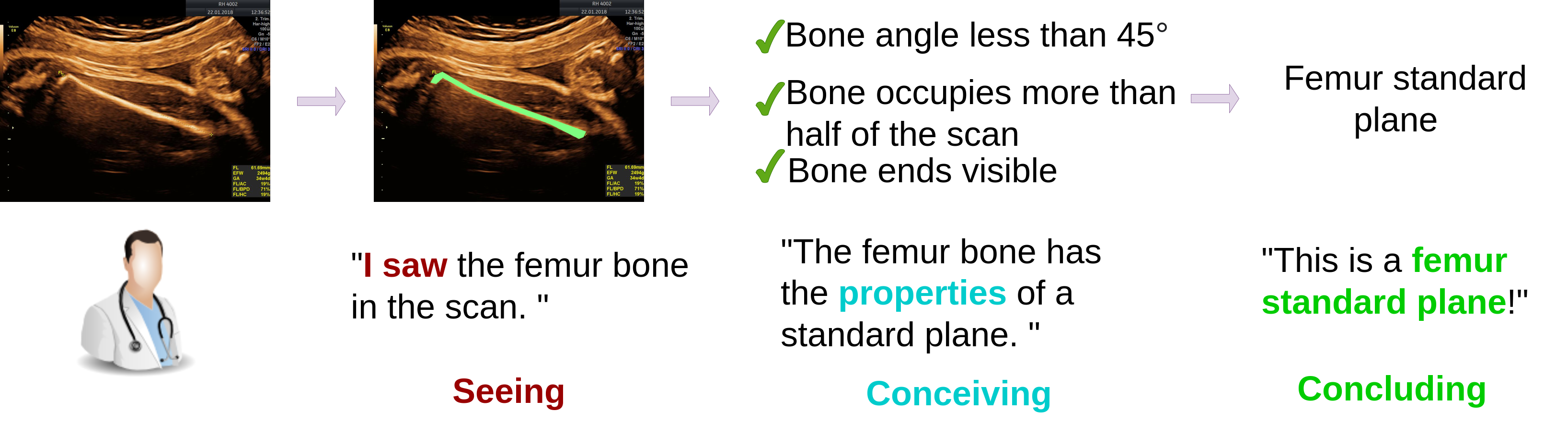}
  \caption{The process of a specialist recognizing a femur standard plane. The concepts in the "Conceiving" stage are taken from~\citep{salomon2019isuog}. }
  \label{fig:caption}
\end{figure}

Explainable US image analysis has so far focused on post-hoc saliency maps~\citep{baumgartner2017sononet}, attention mechanisms~\citep{schlemper2019attention,cai2020spatio} or anatomical structure detection~\citep{he2024fetal,lin2019multi,wu2017fuiqa}. These methods, while providing visual clues of the model decision, often lack faithful and holistic explanations.  We improve on this by introducing a \emph{progressive} CBM $x \xmapsto{g} s \xmapsto{l} c \xmapsto{f} y$ which predicts, first, segmentation concepts $s$, then property concepts $c$, and finally the prediction $y$. This design is holistic, ensuring that explanations are comprehensive and linked to predictions. The design is, however, also designed based on how an experienced sonographer would recognize a high-quality femur standard plane from thousands of ultrasound scans. Fig.~\ref{fig:caption} illustrates the standard procedure~\citep{salomon2019isuog} of first seeking specific organs to have a rough assessment of the image, and subsequently, based on the concepts from their knowledge, making a decision. In other words, sonographers \textit{\textbf{see}} organs and regions, \textit{\textbf{conceive}} their properties according to their knowledge, and finally \textit{\textbf{conclude}} on the content of the image. We utilize these steps of the clinical procedure to rethink concept learning in CBMs as a sequence of  \textit{\textbf{``seeing"}}, \textit{\textbf{``conceiving"}}, and \textit{\textbf{``concluding"}}. This model design is motivated  by expected performance boosts, but also by being directly adaptable to a typical clinical workflow.%, we rethink concept learning in CBMs. %As illustrated in Fig.~\ref{fig:caption}, the femur bone visualization, the expert-specified standard plane requirements, and even the category "femur standard plane", can be considered a progression of concepts with increasing abstraction. Here, abstraction means the level of difficulty for a non-expert to understand. To this end, the visual understanding task is a gradual abstraction process of concepts from redundant information. This is also supported by early post-hoc studies~\citep{zeiler2014visualizing, yosinski2015understanding}, which reveal that shallower layers in a neural network tend to learn simpler concepts, while the higher layers prefer abstract ones. 

\subsubsection{Avoiding unintended harm from information leakage.} Recent results~\citep{margeloiu2021concept, mahinpei2021promises} suggest problems hindering CBMs from being widely trusted and employed in practice: Sequential or joint training of CBMs, where $c$ is represented via soft class probabilities rather than thresholded predictions, is crucial to avoid performance loss, but comes with risk of information leakage from $y$ to the mapping $g$ during the sequential and joint training~\citep{margeloiu2021concept, mahinpei2021promises,margeloiu2021concept,raman2024concept}. This allows CBMs to encode information not only in whether or not a concept is present, but also in its probabilities. As a result, the concepts could carry more information than a user would intuitively realize. The explanations are hereby not faithful. Independent training of CBMs, where concepts $c$ are thresholded before they are passed to the classifier $f$, eliminates this information leakage, but typically come with a noticeable loss of performance. As a result, high-performing CBMs come with risk of being less transparent than they seem: The mapping $x \mapsto c$ is a black-box model, which could encode more information than what meets the eye. This, again, could lead to decreased explanation robustness and consistency, potentially harming the user's trust.- Our P-CBM design minimizes the risk of leakage by using organ segmentations as the first level of concepts, thereby limiting the model's ability to learn from unwanted information.

\subsubsection{Case study in 3rd-trimester growth scans.} Our case study focuses on feedback support for clinicians performing a standard 3rd-trimester growth scan. These scans are challenging because the increased fetus size and reduced amniotic fluid imply poor image quality~\citep{parikh2019third}. However, due to a shortage of trained sonographers, these scans are often performed by less experienced clinicians~\citep{edvardsson2016physicians,recker2021point}. Improving the quality of these scans could therefore have substantial clinical impact. Our model aims to assist these clinicians at multiple stages: Guidance towards improved image quality via visual segmentation concepts while scanning; quality assessment of selected planes using the model predictions; detailed feedback on the quality of the completed scan via property concepts after finishing the scan. 

\subsubsection{Our key contributions are as follows: }
\begin{itemize}
    \item We propose Progressive Concept Bottleneck Models (P-CBMs), which provide both semantic and property explanations aligned with the ISUOG criteria. Our model provides holistic explanations to the clinicians, that show how predictions depend directly on the ISUOG criteria, emphasizing the model's clinical applicability. To our knowledge, we are the first to incorporate all the ISUOG criteria into an explainable model.
    \item The P-CBM architecture, which focuses the model's attention near organs that are known to be relevant for the image quality assessment task, comes with a reduced risk of \emph{leakage}, which is crucial for the faithfulness of the model's explanations.
    \item We formulate the fetal US quality assessment as a challenging but realistic fine-grained image recognition task. We demonstrate how the resulting P-CBM exhibits a boost in performance over non-explainable models, as well as strong generalizability across external datasets, which can likely be attributed to the model's design to avoid leakage as well as to the comprehensiveness of the concepts.
\end{itemize}

\section{Related work}
\label{sec:relatedwork}
\subsection{Fetal ultrasound quality assessment }
Over the past decade, DNNs have increasingly been employed to automate the selection of SPs in fetal US imaging, treating it as an image recognition challenge. Studies such as~\cite{yu2017deep} and~\cite{qu2020standard} have leveraged state-of-the-art DNNs for the detection of fetal SPs.~\cite{burgos2020evaluation} suggest DenseNet has the best performance in classifying six different planes.~\cite{montero2021generative} enhanced classification performance by employing image augmentation with generative models.~\cite{dapueto2024knowledge} achieved real-time performance improvements by distilling knowledge from a larger model into a more compact one.~\cite{sendra2023generalisability} validated the effectiveness of transfer learning by demonstrating the generalizability of DNNs trained with images from Spain and Denmark on datasets from Africa.

An alternative approach conceptualizes the problem as the detection of anatomical structures. According to the ISUOG criteria, the identification of specific anatomical features can improve the explainability of model decisions.~\cite{wu2017fuiqa}employed a localization network to identify the fetal abdomen, verifying the presence of key structures like the stomach bubble and umbilical vein using a classification network.~\cite{lin2019multi}, ~\cite{guo2022fetal}, and~\cite{taksoee2024ai} engaged in multi-task learning, simultaneously localizing anatomical structures and recognizing SPs. Although multi-task learning enriches the feature space, it often lacks causal explanations, meaning the classification outcomes may not directly depend on the detection results.~\cite{slimani2023fetal} integrated segmentation and classification models across different anatomies, showing promising results with rule-based approaches; however, their method does not fully address all ISUOG criteria. Baumgartner et al.~\cite{baumgartner2017sononet} introduced SonoNet, which localizes organs in real time using a weakly-supervised approach.~\cite{schlemper2019attention} implemented a self-gated soft-attention mechanism in a DNN to identify 13 different standard planes, providing visual insights into the decision-making process. Nonetheless, these weakly-supervised methods often fall short in precisely localizing specific organs.

In contrast to these methods, in this paper, we encode the full ISUOG standards into our P-CBM, which boosts both interpretability and model performance. Our model ensures the prediction from each stage has a faithful ground from the previous stage. In addition, our method precisely discriminates between SPs and NSPs of the same anatomy, instead of the common setting of SPs against `background' images. Our task is more challenging but is closer to the real-world ultrasound screenings.  

\subsection{Model explainability}
\subsubsection{Explaining with pixel attributions.}
Pixel attributions, also known as visual-spatial selection attention, allow humans to selectively mask out redundant information in a cognition task~\citep{lockhofen2021neurochemistry}. Explaining with pixel attributions means understanding an image with regions of interest. Class Activation Mapping~\citep{zhou2016cvpr} and its successors~\citep{selvaraju2017grad, wang2020score} localize the features of the DNNs that are responsible for the classification decision on the image by fusing channel contributions. For the fetal US quality assessment task,~\cite{slimani2023fetal} and  \cite{taksoee2024ai} explain the model decision with segmentation results.  
% Region-based recognition methods~\citep{wei2016mask, huang2020interpretable}, in contrast, locate visual concepts of the input images and obtain corresponding feature vectors, based on which the network makes decisions.

Pazzani et al.~\citep{pazzani2022expert} argue that explanations are what a domain expert would use, and visual explanations can not fulfill the explainability requirement in fetal US quality assessment - given the criteria in Tab.~\ref{tab:isuog}, points such as symmetry and caliper placement can not be easily visualized.   

\subsubsection{Explaining with property concepts.}
Property concepts refer to any abstraction distilled by domain experts, e.g., a color, a shape, or even an idea~\citep{molnar2022}. They are used to describe the typical human-understandable characteristic of an image. Explaining property concepts means \textit{\textbf{``conceiving"}} the property of the image before \textit{\textbf{``concluding"}} the content. Several authors,
~\citep{kim2018interpretability,yuksekgonul2022post,ghosh2023dividing}, provide post-hoc concept scores for black-box models.

Different from these post-hoc methods, Concept bottleneck models~\citep{koh2020concept, chen2020concept,xu2024energy, zarlenga2022concept,kim2023probabilistic} represent concepts with neuron activation, which makes them intrinsically interpretable. CBMs~\citep{koh2020concept} enable human explanation inspection and correction by intervening in neuron activation. The limited number of expert-given concepts as well as the lack of expressivity in the predictor constrain the performance of CBMs. Existing works~\citep{sarkar2022framework,yang2023language} supplement the property concepts by self-supervised learning or language models.

\subsection{Information leakage in concept bottleneck models}
Unfortunately, however, all of the above mentioned models suffer from \emph{information leakage}. Recent works~\citep{margeloiu2021concept,mahinpei2021promises,raman2024concept} have highlighted that information leakage occurs when CBMs inadvertently learn spurious correlations for concept prediction, resulting in degraded model performance and robustness. This leakage arises when CBMs are trained either sequentially or jointly, enabling soft concept probabilities rather than discretized ones. The soft concepts enable encoding extra information from the source image in non-intuitive ways, so that the concept bottleneck can actually encode more information than a user's intuitive interpretation of the concepts would indicate. While independent CBMs mitigate information leakage in bottlenecks, they tend to suffer in performance. Moreover, they are prone to locality leakage—where predicted concepts are based on spurious visual clues in the image, leading to changes in concept prediction when irrelevant features are perturbed~\citep{raman2024concept}. Havasi et al.~\citep{havasi2022addressing} completed the independent CBM concepts with an additional side channel, which, however, can allow unintended information to pass the bottleneck~\citep{havasi2022addressing}.

We propose using progressive bottlenecks to address information leakage at multiple levels. Our model learns concepts with multi-stage, i.e., progressive bottlenecks. Specifically, by grounding the concepts in the ISUOG criteria in the fetal US quality assessment task, providing both pixel attributions and concepts as explanations, we obtain both reduced leakage and clinically relevant explanations. Our experiments on external datasets demonstrate the generalizability of our method.

\subsection{Segmentation for classification}
In this paper, we built a segmentation network to \textit{\textbf{``see"}} the visual concepts. Segmentation as part of a classification network is a common solution in some applications, e.g., region-based fine-grained image recognition~~\citep{he2017mask,huang2020interpretable}, scene recognition~\citep{lopez2020semantic} and defect detection~\citep{bovzivc2021end}.
Huang et al.~\citep{huang2020interpretable} weakly localized different parts of objects via region grouping and predicted fine-grained labels from the region features. 
Semantic-aware scene recognition model (SASceneNet)~\citep{lopez2020semantic} treats segmentation as another modality. It learns features from segmentation and the input image in different branches and fuses them to make a final prediction. 

These segmentation-for-classification methods provide visual explanations~\citep{wei2016mask} but risk information leakage. In this paper, we address the leakage by employing visual explanations as an information bottleneck.

\section{Method}
\label{sec:method}
We consider the fetal ultrasound quality assessment task where the input image $x \in {\mathbb{R}}^{h\times w \times m}$ with $m$ channels, height $h$, and width $w$, induces a predicted target $y$. A black box model would learn a direct mapping predicting $x \mapsto y$, which is non-transparent and thus not explainable.

\begin{figure}[t]
  \centering
  \includegraphics[width=\linewidth]{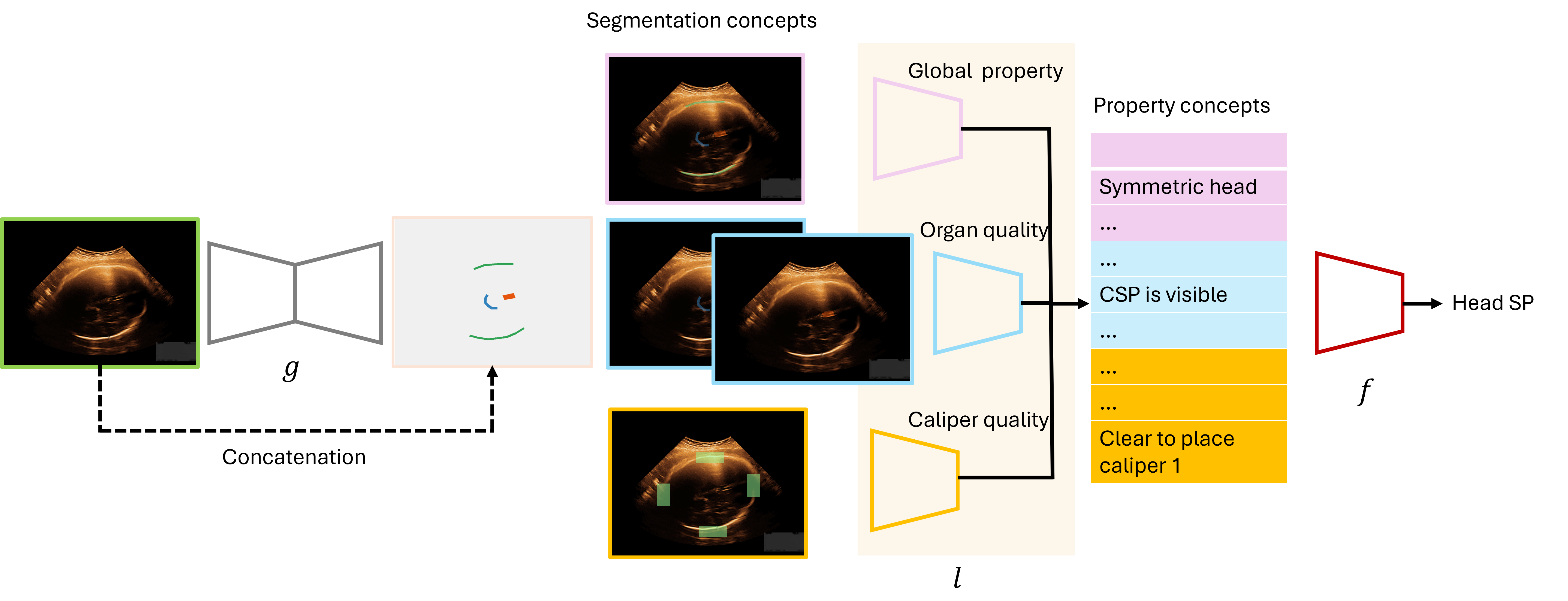}
  \caption{Illustration of the network architecture of the proposed P-CBM.}
  \label{fig:network}
\end{figure}

The fetal US quality assessment task is essentially a hierarchical abstraction process of concepts. Inspired by the heuristic procedure \textit{\textbf{``seeing"}}, \textit{\textbf{``conceiving"}} and \textit{\textbf{``concluding"}}, we propose Progressive Concept Bottleneck Models, which consist of three stages corresponding to the three steps in the procedure. Fig.~\ref{fig:network} demonstrates the architecture of P-CBM. We represent P-CBM as a mapping $f(l(g(x)))$, where $g$ \textit{\textbf{``saw"}} segmentation concepts from $x$, $l$ \textit{\textbf{``conceived"}} property concepts from the segmentation, and $f$ \textit{\textbf{``concluded"}} the prediction from property. To eliminate information leakage~\citep{margeloiu2021concept, mahinpei2021promises}% and to enable concept intervention
, the three stages are trained independently. This means that at each stage, ground truth targets from the previous stage are used for training, whereas predicted targets are used for testing. We introduce the $g$, $l$, and $f$ in detail respectively in section~\ref{subsec:stage1}, section~\ref{subsec:stage2} and section~\ref{subsec:stage3}. 

\subsection{``Seeing" with an observer: Visually intuitive explanations while scanning}
\label{subsec:stage1}
We build an observer network to learn the mapping $g$, which \textit{\textbf{"sees"}} segmentation concepts from the input image $x$. These concepts are segmented regions of interest in the input image $x$, i.e., organs that need to be present in a standard plane. We first trained a segmentation convolutional neural network (CNN) to replicate specialist-annotated segmentation masks $s \in {\mathbb{R}}^{h\times w\times n}$, where $n$ is the number of segmentation concepts. During inference, the observer predicts a segmentation map $p(s | x) \in {\mathbb{R}^{h \times w \times n}}$. We call the output of the observer network the \emph{segmentation bottleneck}, where the model provides visual explanations to users.

\subsubsection{`Soft' and `hard' concepts: Retaining performance while avoiding leakage.}
We integrated the predicted confidence from the observer instead of using hard-coded segmentation predictions. It's important to emphasize that this modification does not result in information leakage. Since different stages of the model are trained independently, neither the perceiver nor the predictor can exploit the bottleneck as proxies to target specific information.

\subsubsection{Objective function: Sensitivity to curvilinear structures.}
We trained a DTU-Net~\citep{lin2023dtu} with an ImageNet~\citep{deng2009imagenet} pre-trained RegNetY-1.6GF~\citep{radosavovic2020designing} backbone as the observer network. The choice of DTU-Net was motivated by its effectiveness in capturing curvilinear structures, frequently encountered in fetal ultrasound images. The objective function remains consistent with the original paper, incorporating terms controlling topology correctness and employing the dice focal loss.

\subsection{``Conceiving" with a perceiver: ISUOG criteria as explanations}
\label{subsec:stage2}
We propose a perceiver network to learn the mapping $l$ from segmentation concepts to high-level property concepts, imitating the process of \textit{\textbf{``conceiving"}} characteristics of seen objects/regions, and associating it with more abstract expert knowledge. In our application, we define the property concepts to be the ISUOG standards~\citep{salomon2019isuog} for the 3rd-trimester standard planes. Based on the prior knowledge, we first divide the property concepts into three categories: global property concepts that describe the attributes of the full image, i.e., the symmetry or the magnification of the scan; organ quality concepts that are related to the visibility of a specific organ; and caliper quality concepts associated with the visibility of the region of interest for a specific caliper. According to the ISUOG standards, these concepts do not only rely on the shape but also the position and color of the objects. Specifically, the global property concepts are based on the interaction of multiple objects, while the organ quality and caliper quality concepts describe the property of a single region of interest. For these reasons, we concatenate the input image with the segmentation concepts as the input of the perceiver network, to fuse texture information. We create three CNNs for the three types of property concepts respectively, and construct the input for each CNN as follows:  

\subsubsection{Organ quality concepts.}
Each organ quality concept is affiliated with a segmented organ in the predicted segmentation map $p(s_i | x) \in {\mathbb{R}^{h \times w}}$. Since the segmentation map does not include texture information, which is important for deriving object properties, we construct the network's input by concatenating the segmentation map with the original image $x$. As a result, the input related to the $i$-th segmentation concept $\bar s_i \in {\mathbb{R}^{h \times w\times (m+1)}}$ is defined as:
\begin{equation}
    \bar s_i = p(s_i | x) \oplus x 
\label{eq1}
\end{equation}
where $\oplus$ means concatenation.

\subsubsection{Caliper quality concepts.}
\begin{figure}[t]
\centering
\includegraphics[width=0.7\linewidth]{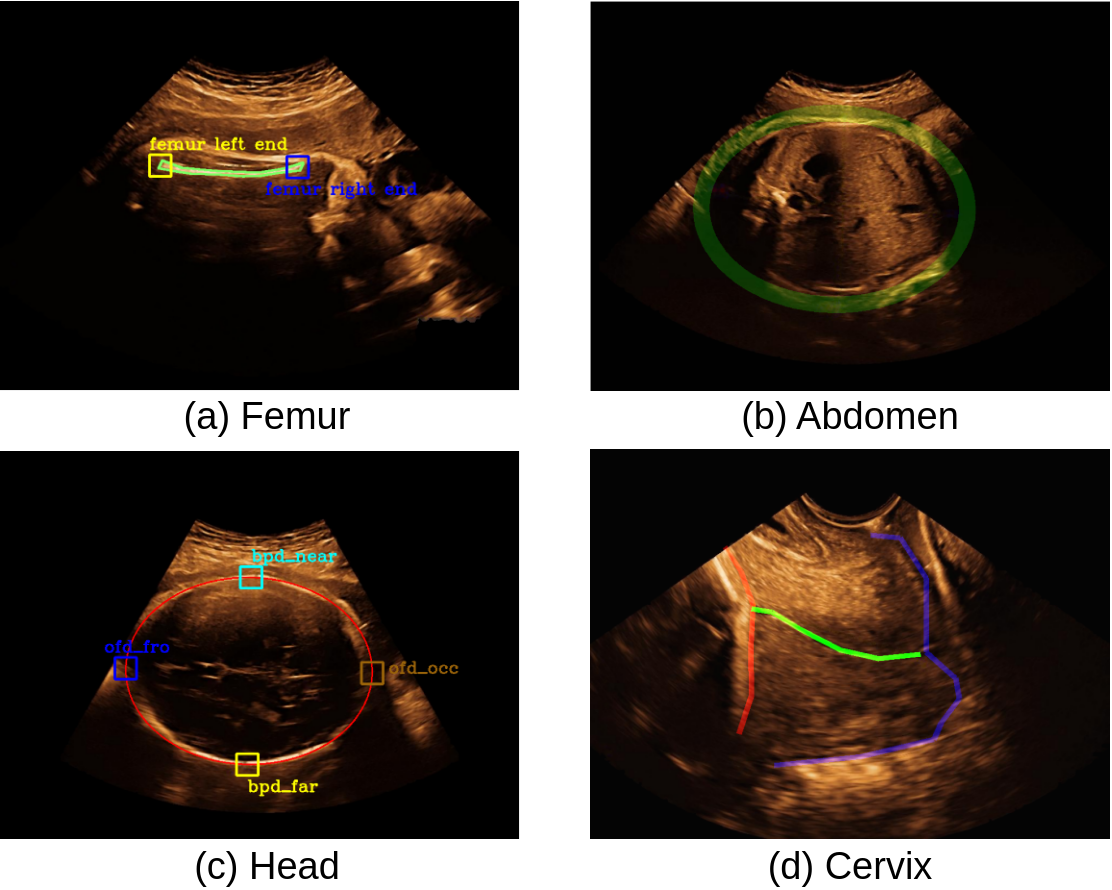}
\caption{Examples of the regions of interest associated with different property concepts. (a), (b), (c), and (d) present the regions for the femur, abdomen, head, and cervix anatomy respectively.}
\label{fig:caliperroi}
\end{figure}

Caliper quality concepts describe the possibility of a region in the scan for caliper placement. Although also relying on a single region of interest, these concepts are not the direct property of the segmentation concepts (organs). Instead, we compute a region of interest based on the segmentation. For example, for the calipers measuring the length of the femur bone, the region of interest is defined as the $40\times40$ image patches on both ends of the segmented femur, where the clinicians usually put the calipers. The input of the CNN is the concatenation of the image $x$ and a binary mask of the corresponding region of interest. An example is illustrated in Fig.~\ref{fig:caliperroi}. 

\subsubsection{Global property concepts.}
Since the global property concepts rely on multiple objects, we concatenate the image $x$ with the segmentation maps $p(s_i | x)$ for all $i=1, \ldots, n$. 

Following previous CBMs, we assume different property concepts to be independent of each other. Hence, the CNNs in the perceiver are trained in a multi-task learning manner, with each output neuron aligned with one property concept. Given $d$ the number of global property concepts, the CNN learning global property concept has $m+n$ input channels and $d$ output channels. The CNN for organ quality concepts has $m+1$ input channels and $\mu$ output channels, given the number of concepts $\mu$. It sequentially runs on different segmentation concepts and predicts the corresponding logit in the output. The CNN for caliper quality concepts works in the same manner as the organ quality CNN but on the generated regions of interest. The outputs of the three CNNs are concatenated as the predicted property concept, i.e., the output of the perceiver.  

\subsubsection{Auto-correction based on segmentation.}
We employ the association of segmentation and property concepts to correct the perceiver-predicted concepts. That is when the object is not appearing in the segmentation prediction, the corresponding property concepts are set to 0. For instance, given the case in Fig.~\ref{fig:caption}, if the femur bone is not presented in the image, all the concepts will be filled with 0 regardless of the original model prediction.       

\subsubsection{Objective function.}
To train the perceiver, three instances of ResNet-18~\citep{he2016deep} pre-trained on ImageNet were employed, each dedicated to learning global property concepts, organ quality concepts, and caliper quality concepts, respectively. This choice was made due to observations indicating that models initialized with pre-trained ImageNet weights exhibited faster convergence compared to those initialized randomly. Adjustments were made to the first convolution layer and the fully connected layer to accommodate the size of the concepts. The property concepts encompass both categorical (e.g., plane symmetry) and numerical (e.g., organ qualities) characteristics. For categorical concepts, the objective function is computed using a weighted binary cross-entropy loss $\mathcal{L}_{bce}$, with weights derived from class density. For numerical concepts, the objective function comprises a Huber loss $\mathcal{L}_{huber}$ for regression control and a Hinge loss $\mathcal{L}_{hinge}$ for rank preservation. Empirically, the objective function of the perceiver can be expressed as follows:
\begin{equation}
    \mathcal{L}_{perceiver} = \mathcal{L}_{bce} + \mathcal{L}_{huber} + 0.1\cdot \mathcal{L}_{hinge}
\end{equation}

\subsection{"Concluding" with a predictor: Is it a standard plane?}
\label{subsec:stage3}
The final predictor \textit{\textbf{"concludes"}} targets from property concepts. The predictor learns $f \colon c \mapsto y$ from annotated concepts when training, and predicts $f(l(g(x)))$ from predicted concepts at test time.

The predictor is an MLP with a hidden layer containing 256 neurons. The hidden layer is equipped with batch normalization and ReLU activation. The objective function is a cross-entropy loss.  

\section{Experiments}
\label{sec:experiments}
We trained and validated the performance of our model on an excerpt from an anonymous national fetal ultrasound database.
% the Danish national fetal ultrasound database.

\subsection{Data preparation}
We extracted n = 3,775 images of 241 subjects from a standard 3rd-trimester growth screening ultrasound examination. These images encompassed both standardized planes (SPs) and non-standardized planes (NSPs) from four specific anatomical regions: the head, femur, abdomen, and maternal cervix. These standardized planes are routinely captured to document fetal biometric parameters relevant to fetal growth and the risk of preterm birth.

\begin{table}[t]
\centering
\begin{minipage}{0.35\textwidth}
    \centering
  \caption{Overview of dataset and division across classes.}
  \label{tab:data}
    \resizebox{\textwidth}{!}{
  \begin{tabular}{ll|ll|ll|ll}
    \toprule    \multicolumn{2}{c|}{femur} & \multicolumn{2}{c|}{abdomen}& \multicolumn{2}{c|}{head} & \multicolumn{2}{c}{cervix}\\
    \midrule
    SP & NSP & SP & NSP & SP & NSP & SP & NSP \\
    \midrule
    346 & 422 & 221 & 784 & 172 & 1026 & 452 & 352\\
    % \bottomrule
    \hline
  \end{tabular}
  }
\end{minipage}
\hfill
\begin{minipage}{0.6\textwidth}
    \centering
  \caption{Concept classification performance of different models on the ultrasound dataset. }
  %`RMSE' means the root-mean-square deviation of the learned concepts to ground truth. `COA' means the binary classification accuracy of concepts. `MRMSE' means the average root-mean-square deviation of the learned concepts to ground truth over different values. `MCOA' means the category-average classification accuracy of concepts.}
  \label{tab:concept_performance}
  \resizebox{\textwidth}{!}{
  \begin{tabular}{lcccc}
    \toprule
    Method & RMSE & COA (\%) & MRMSE & MCOA (\%)  \\
    \midrule
    CBM & $2.08\pm0.024$ & $81.84\pm0.30$ & $3.54\pm0.039$ &$58.80\pm 0.78$ \\
    MTL & $1.76\pm0.178$ & $97.76\pm0.21$ & $4.57\pm0.074$ &$97.75\pm0.32$ \\
    % \midrule
    \rowcolor[RGB]{222,222,222}
    P-CBM (ours) & $\textbf{0.42}\pm0.028$ & $\textbf{98.23}\pm0.15$ & $\textbf{1.58}\pm0.067$ & $\textbf{98.07}\pm0.25$\\
    % \bottomrule
    \hline
  \end{tabular}
  }
\end{minipage}
\end{table}

\begin{figure}[t]
\centering
\includegraphics[width=0.9\linewidth]{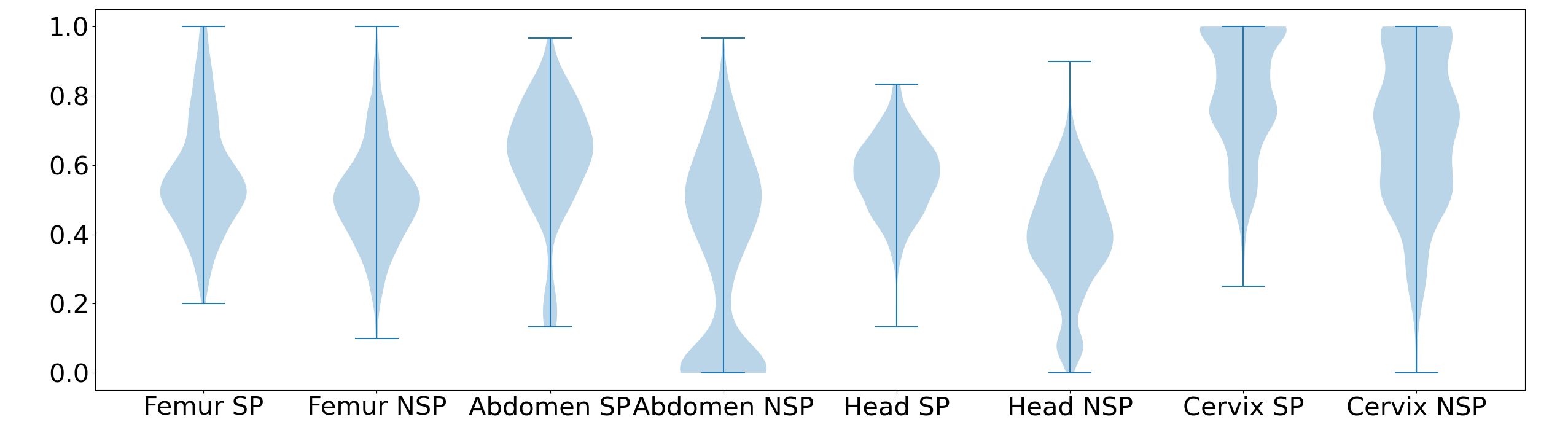}
\caption{Image quality distribution of different classes.}
\label{fig:violin}
\end{figure}

The dataset statistics are detailed in Table~\ref{tab:data}. We note the SP/NSP imbalance in our dataset is the clinical
reality, and we want to validate our model in realistic data distributions. In addition, we prioritized a large test set to stably estimate performance. Thereby, for each test, the experiment was repeated 10 times resampling training (50\%), validation (10\%), and test (40\%) sets, divided on the subject level to avoid subject overlap between splits.

The recorded planes were annotated\footnote{https://github.com/wkentaro/labelme} by an expert (MD, Ph.D. fellow in Fetal Medicine). The images were divided into eight classes consisting of the four anatomies along with an SP/NSP classification; which was done according to the ISUOG criteria summarized in Table~\ref{tab:isuog}. NSPs are the scans not satisfy the criteria and those that do not satisfy the criteria are SPs. The relevant anatomical structures were manually outlined as segmentation concepts.

There are 14 segmentation concepts and 27 related property concepts (shown in Fig.~\ref{fig:concept_performance}). The property concepts were defined based on the ISUOG criteria. The Concepts were either binary (true/false) representing correct angle, image magnification (size of the object), and symmetry or a scalar quality score of 0-10. Quality scores assess the visual quality of anatomical structures and caliper locations (0 = not visible, 10 = excellent visualization, 1-2 means a small part of the organ is visible, and so on). We will release the annotation
protocol and the data examples for transparency and reproducibility. For 85 images, the quality score (scale 1-10) was re-annotated with a more-than-one-month interval; the average score difference was 1.345, which indicates fairly stable annotation.

All images were annotated by one annotator but selected low- and high-quality annotated images were sent to two fetal medical experts to ensure continued agreement in the annotation process and to minimize potential annotation bias. Images with huge disagreement or need for discussion were presented to a panel of fetal medical experts. The panel helped the annotator decide on difficult samples. 

Fig.~\ref{fig:violin} shows the image quality distribution of different classes. The figure demonstrates that in each class, the image quality varies a lot. Image quality is determined by the average annotated organ quality within each image.

\subsection{Implementation details}
In line with CBMs, each image's property concept label $c$ is a 27-dimensional vector. In this vector, each element represents a concept, with a padding of 0 for concepts not applicable to the image. Images were resized to $224\times 288$ and pixel intensity normalized to $[-1, 1]$. We removed the texts and calipers in each image with~\citep{mikolaj2023removing}.

The observer, perceiver, and predictor were trained independently using the AdamW optimizer for 200, 100, and 50 epochs, respectively. A weight decay of 1e-6 was applied. The initial learning rate was set to 1e-4, with a reduction by a factor of 0.1 if the validation loss plateaued for 10 epochs. A batch size of 32 was utilized for all three stages. Models with the best performance on the validation set were selected for evaluation on the test set. To address class imbalance, oversampling was implemented for images from categories "femur SP," "abdomen SP," "head SP," and "cervix NSP" during training. The code was developed using Python 3.7.7 and PyTorch 1.12.1, and experiments were conducted on an Almalinux 8.7 platform with a Quadro RTX A6000 GPU.

The ultrasound images in our dataset were extracted from DICOM images saved by the operators. These images contain some calipers placed by the operator during screening. These calipers act as confounding information for our caliper quality concept learning, which measures to which degree on the image the operator can place a specific caliper - the existing calipers on the image provide model direct hints. To eliminate this confounding information, we captured the existing calipers by thresholding in the HSV space, and then inpainted them with~\cite{mikolaj2023removing}. Texts about the scan information were removed, either. 
% An example is given in Fig.~\ref{fig:caliper}.    

% \begin{figure}[h]
%     \centering
%     \subfigure[Raw image.]{
%     \begin{minipage}[t]{0.45\linewidth}
%     \centering
%     \includegraphics[width=0.8\linewidth]{figures/before.png}
%     \end{minipage}
%     }
%     \subfigure[Clean image.]{
%     \begin{minipage}[t]{0.45\linewidth}
%     \centering
%     \includegraphics[width=0.77\linewidth]{figures/after.png}
%     \end{minipage}
%     }
%     \caption{Examples of confounding information removal. The image is cropped to keep anonymity. Zoom in for details.}
%     \label{fig:caliper}
% \end{figure}

\begin{table}[t]
\centering
\begin{minipage}{0.47\textwidth}
    \centering
  \caption{Classification performance of different models on the ultrasound dataset. }
  % `OA' means the overall classification accuracy over instances and `MA' means the average accuracy over categories. `MCC' means Matthews correlation coefficient. }
  \label{tab:performance}
    \resizebox{\textwidth}{!}{
    \begin{tabular}{lccc}
    \toprule
    Method & OA (\%) & MA (\%) & MCC (\%) \\
    \midrule
    Standard Model & $62.71\pm5.67$ & $61.64\pm1.68$ & $57.62\pm5.12$\\
    SonoNet-32~\citep{baumgartner2017sononet} & $68.83\pm3.13$ & $62.73\pm0.81$ & $63.81\pm3.02$\\
    \midrule
    SASceneNet-18\citep{lopez2020semantic} & $73.71\pm2.86$ & $68.19\pm1.88$ &$69.60\pm3.19$ \\
    MTL~\citep{he2024fetal} & $73.75\pm2.75$ & $70.39\pm2.19$ &$69.90\pm2.88$ \\
    \midrule
    CBM~\citep{koh2020concept} & $72.24\pm1.99$ & $72.22\pm0.45$ & $68.24\pm1.92$ \\
    % \midrule
    \rowcolor[RGB]{222,222,222}
    P-CBM (ours) & $\textbf{80.22}\pm2.15$ & $\textbf{78.18}\pm1.95$ & $\textbf{77.17}\pm2.37$\\
    % \bottomrule
    \hline
  \end{tabular}
  }
\end{minipage}
\hfill
\begin{minipage}{0.47\textwidth}
    \centering
  \caption{Classification overall accuracy of different models on external datasets. }
  %`FPDB-Head' means the head SP/NSP classification task in the FPDB dataset}
  \label{tab:externaldata}
    \resizebox{\textwidth}{!}{\begin{tabular}{lccc}
    \toprule
    \multirow{2}{*}{Method} & \multicolumn{3}{c}{Datasets}\\
    \cline{2-4}
     ~  & FPDB & FPDB-Head & Africa\\
    \midrule
    Standard Model & $76.46\pm5.44$ & $43.52\pm1.57$ & $84.64\pm6.73$\\
    SonoNet-32 & $85.88\pm3.74$ & $43.73\pm1.91$ & $79.71\pm2.83$\\
    \midrule
    SASceneNet-18& $91.39\pm2.25$ & $43.73\pm1.91$ & $85.49\pm4.77$\\
    MTL & $\textbf{96.52}\pm1.09$ & $44.88\pm3.00$ & $94.85\pm0.86$\\
    \midrule
    CBM & $91.26\pm0.49$ & $44.95\pm0.62$ & $95.44\pm0.68$\\
    % \midrule
        \rowcolor[RGB]{222,222,222}
    P-CBM (ours) & $96.00\pm0.76$ & $\textbf{58.54}\pm3.10$ & $\textbf{96.35}\pm0.88$\\
    % \bottomrule
    \hline
  \end{tabular}}
\end{minipage}
\end{table}

\subsection{Comparison with state-of-the-arts}
\subsubsection{Baselines.}
We benchmark our model's performance against several baseline models. The first baseline, referred to as the standard model, shares the same architecture as our P-CBM but is trained end-to-end for classification purposes. Additionally, we include SonoNet-32~\citep{baumgartner2017sononet}, a state-of-the-art network for ultrasound plane classification, and SASceneNet-18~\citep{lopez2020semantic}, which integrates segmentation into a classification model similarly to our P-CBM. Besides these black-box models, we trained a standard CBM as a baseline as well. For a fair comparison, we used the same `observer' for segmentation in SASceneNet-18. The same `perceiver' and `predictor' were used to build CBM. 

Furthermore, we introduced a multi-task learning (MTL) model adapted from~\citep{he2024fetal}. This MTL model performs concept prediction, image classification, and segmentation concurrently across three branches. It utilizes a shared feature encoder, RegNetY-1.6GF, and a segmentation decoder. Additionally, it includes two separate heads for learning property concepts and image categories. All baseline models were trained using identical settings as our P-CBM.

\subsubsection{Quality assessment results.}
We evaluated the model performance on standard plane detection with three metrics: overall classification accuracy over instances (OA), average classification accuracy over categories (MA), and Matthews correlation coefficient (MCC)~\citep{chicco2020advantages}. Given the class imbalance, MA and MCC can better reflect the model performance on underrepresented classes compared to OA. 

As shown in Table~\ref{tab:performance}, our P-CBM demonstrates superior performance across all three metrics compared to the baselines. Specifically, our model outperforms other methods by a margin of at least $7.27\%$ in terms of MCC over 10 splits, indicating enhanced recognition of underrepresented classes. Compared to the CBM model, our P-CBM improves $7.98\%$ in OA. This enhancement can be attributed to our model's ability to not only \textbf{``conceived"} but also \textbf{``saw"} concepts.

\begin{figure*}[t]
\centering
\includegraphics[width=\linewidth]{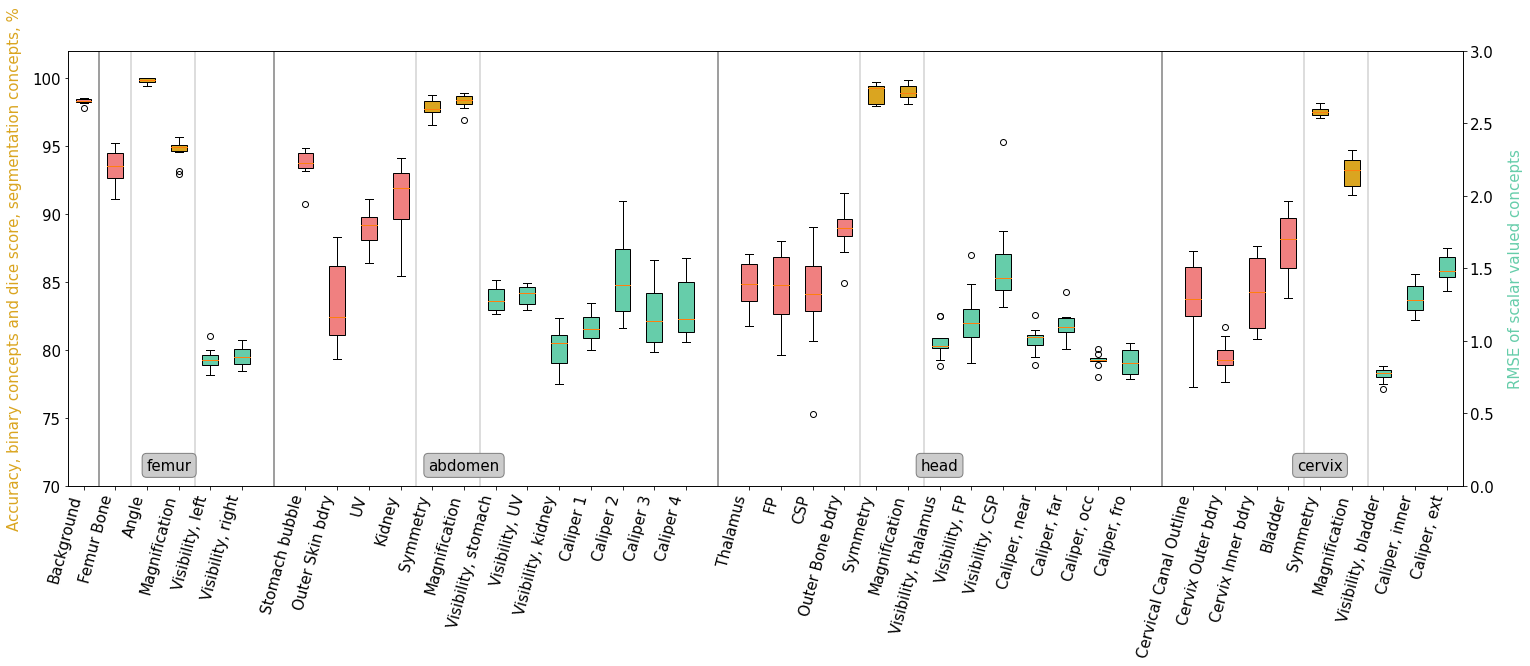}
\caption{Concept validation: Segmentation concepts shown in red refer to dice score values on the left-hand y-axis; these are expressed in $\%$. Binary property concepts (angle, magnification, symmetry) shown in gold refer to values in $\%$ on the left-hand y-axis. Scalar property concepts describing the visibility of organs and caliper locations shown in green refer to values on the right-hand y-axis. Best viewed in color. Here, CSP means cavum septi pellucidi; UV means umbilical vein and FP means Fossa Posterior.}
\label{fig:concept_performance}
\end{figure*}

\subsubsection{Correctness of explanation.}
Our model's performance hinges on two intermediate layers of concepts: segmentation and property. Segmentation concepts are validated using the dice score, illustrated in Fig.~\ref{fig:concept_performance}. Our observer net achieves high dice scores in numerous categories, though performance drops are noticeable in thin and elongated curvilinear structures like the outer skin boundary and thalamus, where lower dice values are expected despite good performance.

On the other hand, property concepts are validated in terms of classification accuracy for binary concepts and root mean squared error (RMSE) for scalar concepts, also shown in Fig.~\ref{fig:concept_performance}. Binary concepts exhibit classification accuracy well above $93\%$, while most scalar concepts achieve an RMSE of 1.5 or below.

Additionally, a comparison across concepts between our model, MTL model, and a standard CBM model is performed, as detailed in Table~\ref{tab:concept_performance}. The property concepts also exhibit imbalance across different categories/values. The table reports the average accuracy of binary concepts over categories (MCOA) and the average RMSE of scalar concepts over ground truth values ranging from 0 to 10 (MRMSE). Results demonstrate that our model outperforms baselines even in underrepresented concept values, indicating its capability to provide accurate explanations to guide sonographers' screening processes.

\subsubsection{Validation on external datasets.}
We demonstrate that our model's performance is transferable to other datasets without the need for any transfer learning. This capability is particularly advantageous for training the model in a clinical center with high-resource settings and then deploying it to a new center with low-resource settings.

We assess our model's performance using two publicly available datasets: the FPDB dataset~\citep{burgos2020evaluation}, containing 9,418 images covering the femur, abdomen, head, and cervix, and the Africa dataset~\citep{sendra2023generalisability}, comprising femur, abdomen, and head images from 120 patients across five African countries. We evaluate the model performance in classifying different anatomical regions. Moreover, the FPDB dataset offers fine-grained annotations for head images. Specifically, the "trans-thalamic" plane within the FPDB dataset corresponds to our definition of head SP, while other head planes represent head NSP. We further benchmark the model's performance on the fine-grained head SP/NSP classification task.

The results are presented in Table~\ref{tab:externaldata}. Our model demonstrates competitive performance in the anatomy classification task on both the FPDB and Africa datasets and achieves outstanding performance on the fine-grained head SP/NSP benchmark. These findings indicate that our model can be effectively transferred to unseen datasets without the need for fine-tuning. It is worth noting that the explanations generated by our model can also be transferred across datasets.

Comparing the performance of SASceneNet-18, MTL, and CBM with that of SonoNet-32, we contend that the introduction of concepts, i.e., prior knowledge and rules, enhances model generalizability. This observation is reasonable because the definition of concepts is more general compared to the implicit knowledge learned by the model from the data.

\subsection{Model analysis}
\subsubsection{Consistency of explanations.}
Ideally, property concepts should be predicted based on relevant parts of the input space~\citep{margeloiu2021concept}. P-CBM, incorporating an additional segmentation bottleneck compared to CBM, facilitates the learning of property concepts from semantically meaningful regions within the input image. This segmentation stage not only provides visual explanations but also promotes accurate concept learning.
\begin{table*}
  \begin{center}
  \caption{Predicted organ quality value in the inpainted images.}
  \label{tab:rol}
    \resizebox{\textwidth}{!}{
  \begin{tabular}{lccccccc}
    \toprule
    Method & Thalamus & Cavum Septi Pellucidi & Stomach Bubble & Umbilical Vein & Bladder & Kidney & Fossa Posterior\\
    \midrule
    CBM & $2.78\pm0.05$& $1.72\pm0.11$& $3.73\pm0.05$& $1.89\pm0.15$ & $1.81\pm0.45$ &  $1.36\pm0.18$ &  $1.33\pm0.37$\\
    MTL & $0.17\pm0.12$& $0.60\pm0.16$& $0.96\pm0.85$& $1.33\pm1.11$ & $0.89\pm0.94$ &  $\textbf{0.02}\pm0.04$ &  $2.57\pm2.03$\\
        \rowcolor[RGB]{222,222,222}
    P-CBM & $\textbf{0.08}\pm0.18$& $\textbf{0.00}\pm0.00$& $\textbf{0.52}\pm0.29$& $\textbf{0.00}\pm0.00$ &  $\textbf{0.17} \pm 0.33$ &$0.44\pm 0.45$ &  $\textbf{0.21}\pm 0.37$ \\
    % \bottomrule
    \hline
  \end{tabular}
  }
  \end{center}
\end{table*}
Explanation consistency, as defined in~\citep{das2020opportunities}, measures the amplification of explanation change when input data changes. To validate this, we employed image inpainting~\citep{telea2004image} to remove specific organs (thalamus, cavum septi pellucidi, stomach bubble, umbilical vein, bladder, kidney, and fossa posterior) from their respective images and replaced them with nearby textures. The inpainting area is the bounding box of the annotated region of interest dilated by 5 pixels, which helps mask out the organ shape information in the image. We then tested CBM, MTL, and our P-CBM on these inpainted images. Each experiment was conducted across all images containing the targeted organ. Ideally, if the model correctly learns concepts from the image, images lacking these organs (e.g., inpainted images) should yield zero predictions for the associated quality concepts. Therefore, we evaluated the model performance based on the value of the associated concept, as shown in Table~\ref{tab:rol}.

Results demonstrate that our P-CBM provides more consistent explanations, particularly evident in cases like cavum septi pellucidi and umbilical vein, where our model consistently predicts zero for all images. This improvement is attributed to the segmentation bottleneck, enabling our model to first identify regions of interest in the image before predicting property concepts. In contrast, CBM lacks this segmentation stage, leading to confounding information within the image. While MTL alleviates the problem to some extent, it still struggles with organs like the fossa posterior.

% Fig.~\ref{fig:example_role_of_seg} shows an example of the consistency experiment. On the top row, the image is predicted to be an abdomen standard plane since the stomach bubble has a quality grade of 7.52 (this is in accordance with the ground truth). However, when we inpainted the stomach bubble from the image on the middle and bottom rows. Our P-CBM (bottom) updated the quality grade to 0, which means that the stomach bubble was no longer visible in the image, while CBM still predicted the quality to be 6.03, which is not correct. This incorrect concept prediction then led to the incorrect plane classification result.

% \begin{figure}
% \centering
% \includegraphics[width=0.9\linewidth]{figures/roleofseg.png}
% \caption{An example showing the role of the segmentation bottleneck. The top row is the inference result of P-CBM on an abdomen standard plane. The quality grade of the stomach bubble on the image is predicted to be 7.52. On the middle and the bottom row, the stomach bubble on the image is inpainted. They show the inference result of P-CBM and CBM respectively.}
% \label{fig:example_role_of_seg}
% \end{figure}

\subsubsection{Test-time intervention.}
Similar to CBM, our P-CBM allows users to inspect and intervene on both concept bottlenecks to correct wrong explanations. In our user case, the intervention is useful in the downstream stage of biometric measurement and estimation of e.g. fetal weight, where clinicians can disagree with the model and document this. The model prediction is updated when the explanations are corrected. We simulated this by replacing predicted concepts with ground truth and recording the change in results.

\begin{figure}[t]
\centering
\includegraphics[width=0.9\linewidth]{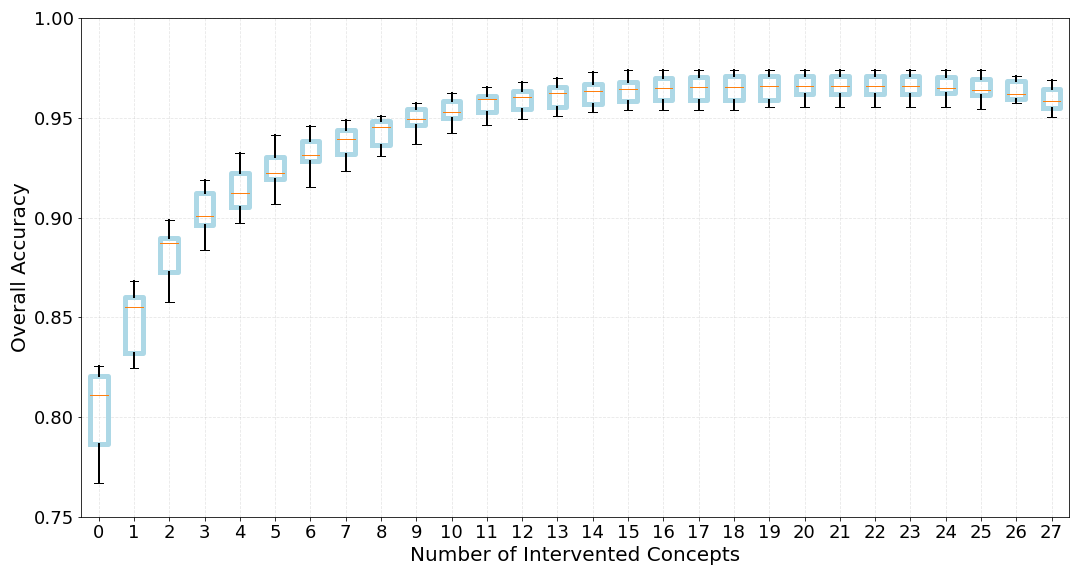}
\caption{Test-time intervention results at the property bottleneck. Better view in color.}
\label{fig:concept_intervention}
\end{figure}

At the property level, we intervened concepts one by one by greedy best-first search~\citep{xie2014jasper} at each split. We reported the tendency of the mean and standard deviation of overall classification accuracy over 10 splits with the increase of intervened concepts in Fig.~\ref{fig:concept_intervention}. We see from the figure that the intervention helps our P-CBM improve the classification performance.

\begin{table}[htbp]
\centering
\begin{minipage}{0.6\textwidth}
    \centering
  \caption{Test-time intervention at the segmentation bottleneck. }
  \label{tab:seminterv}
    \resizebox{\textwidth}{!}{
  \begin{tabular}{lcccc}
    \toprule
    Method & RMSE & COA (\%) & MRMSE & MCOA (\%)  \\
    \midrule
    \textit{w/o} intervention & $0.42\pm0.028$ & $98.23\pm 0.15$ & $1.58\pm0.067$ &$98.07\pm0.25$ \\
    \textit{w} intervention & $\textbf{0.37}\pm0.031$ & $\textbf{98.72}\pm0.07$ & $\textbf{1.54}\pm0.075$ &$\textbf{98.55}\pm0.14$ \\
    \hline
  \end{tabular}
  }
\end{minipage}
\hfill 
\begin{minipage}{0.35\textwidth}
    \centering
  \caption{Ablation studies on different components.}
  \label{tab:ablation}
      \resizebox{\textwidth}{!}{
  \begin{tabular}{ccccc}
    \toprule
    SB & CB & OA (\%) & MA (\%) & MCC (\%) \\
    \midrule
    & \checkmark & $72.24\pm1.99$ & $72.22\pm0.45$ & $68.24\pm1.92$ \\
    \checkmark & & $75.71\pm3.02$ & $73.11\pm1.50$ & $72.11\pm3.12$\\
\rowcolor[RGB]{222,222,222}
    \checkmark & \checkmark  &$\textbf{80.22}\pm2.15$ & $\textbf{78.18}\pm1.95$ & $\textbf{77.17}\pm2.37$\\
    % \bottomrule
    \hline
  \end{tabular}
  }
\end{minipage}
\end{table}

\begin{figure}[t]
    \centering
    \includegraphics[width=0.9\linewidth]{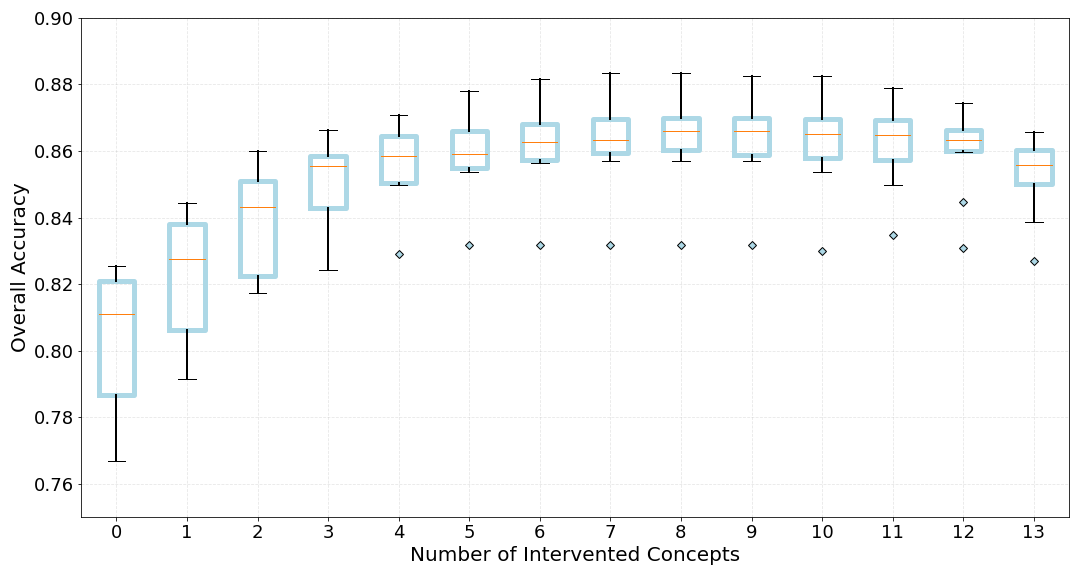}
    \caption{Test-time intervention results at the segmentation bottleneck.}
    \label{fig:interv}
\end{figure}

Our P-CBM allows human intervention also at the segmentation bottleneck. That is when the model is applied to downstream tasks such as biometric measurement of the fetus, the user can correct the wrong segmentation explanations. The model re-inferences on the corrected concepts and gives an updated prediction. We intervened in segmentation concepts one by one
by greedy best-first search at each data split and reported the result in Fig.~\ref{fig:interv}. Quantitive results with all concepts intervented are in Table~\ref{tab:seminterv}. The intervention improved the predictive performance of property concepts and thus improved the classification result. This is because our perceiver modeled a relationship between the segmentation concepts and the property concepts.

\begin{figure}
    \centering
    \includegraphics[width=0.9\linewidth]{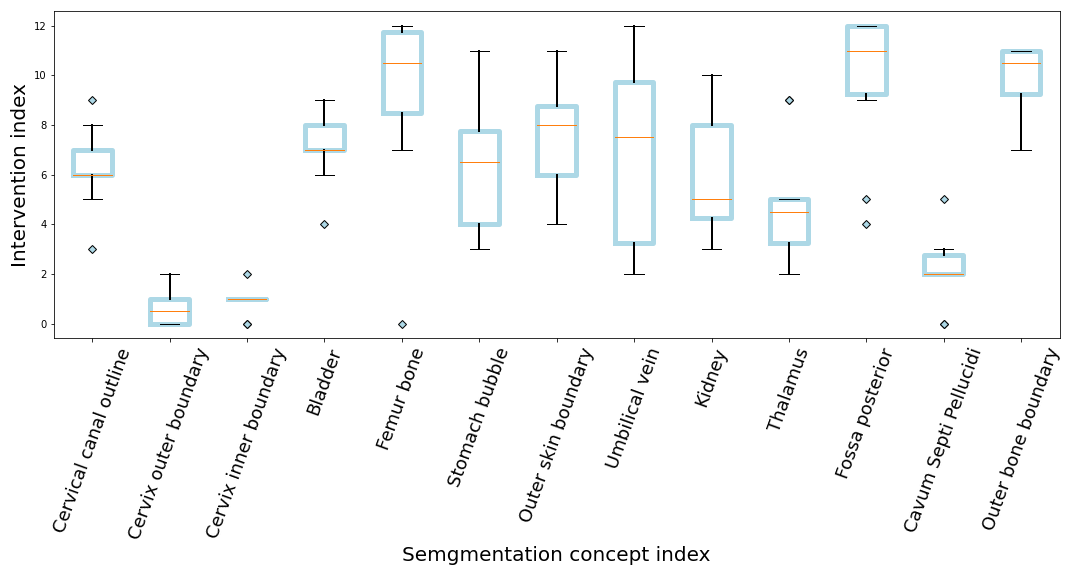}
    \caption{Intervention index of segmentation concepts.}
    \label{fig:index}
\end{figure}

We counted the index of each segmentation concept being selected in the sequence during the intervention. That is, the first concept picked by the greedy search algorithms has an index of 0, and the second has an index of 1, etc. The index of each concept was averaged over 10 data splits. A lower index means a higher impact when the concept is corrected in Fig.~\ref{fig:interv}. The index of each segmentation concept is shown in Fig.~\ref{fig:index}. According to the figure, the intervention on `Cervix outer boundary' and `Cervix inner boundary' helps improve the model performance more than other concepts. This is related to the lowest segmentation performance of the concept, in line with Fig.~\ref{fig:concept_performance}.

% Fig.~\ref{fig:example_interv} shows an example of successful human-model communication in our P-CBM. On the top row of the figure, our P-CBM made a wrong prediction, since the observer failed to recognize the kidney (the yellow area on the middle-row segmentation). Human experts corrected the visual explanation by changing the intermediate segmentation output on the middle row, which also altered the predicted kidney quality concept from 0 to 0.98, and affected the classification result. On the bottom row of the figure, we show the property-level intervention, the same as that in CBM, which ignores the segmentation bottleneck and directly communicates with the property bottleneck. This intervention also corrects the prediction successfully. 

% \begin{figure}
% \centering
% \includegraphics[width=0.9\linewidth]{figures/intervention.png}
% \caption{An example of successful human intervention at different levels is the proposed P-CBM. The top row is a wrong prediction from our model because the model does not recognize the kidney in the image. The middle and the bottom rows illustrate the human intervention at the segmentation and property levels respectively. The number in the figure stands for the predicted kidney quality concept.}
% \label{fig:example_interv}
% \end{figure} 

\subsubsection{Ablations on the information fusion in the perceiver.}
\begin{table*}[t]
  \begin{center}
      \caption{Ablation study results on the information fusion in the perceiver.}
  \label{tab:fusion}
  \smallskip\noindent
    \resizebox{\textwidth}{!}{
  \begin{tabular}{cccccccccc}
    \toprule
    Shape & Texture & Soft concepts & OA (\%) & MA (\%) & MCC (\%) & RMSE & MRMSE & COA(\%) & MCOA(\%) \\
    \midrule
    &\checkmark & & $74.93\pm2.10$ & $69.42\pm2.17$ & $71.46\pm2.22$ & $0.47\pm0.04$ & $1.92\pm0.09$ & $98.21\pm0.15$ & $98.05\pm0.25$\\
    \checkmark& & & $69.94\pm3.08$ & $70.23\pm1.43$ & $66.36\pm2.92$ & $0.56\pm0.06$ & $2.10\pm0.23$ & $97.98\pm0.17$ & $97.54\pm0.32$\\
    \checkmark& & \checkmark& $70.28\pm3.40$ & $70.37\pm1.55$ & $66.72\pm3.27$ & $0.55\pm0.07$ & $2.04\pm0.28$ & $97.98\pm0.17$ & $97.54\pm0.32$\\
    \checkmark& \checkmark &  & $78.42\pm2.20$ & $77.43\pm2.15$ & $75.22\pm2.40$ & $0.45\pm0.03$ & $1.68\pm0.77$ & $98.23\pm0.15$ & $98.08\pm0.25$\\
    % \midrule
    \hline
        \rowcolor[RGB]{222,222,222}
    \checkmark& \checkmark & \checkmark & $\textbf{80.22}\pm2.15$ & $\textbf{78.18}\pm1.95$ & $\textbf{77.17}\pm2.37$ & $\textbf{0.42}\pm0.028$ & $\textbf{1.58}\pm0.067$ & $\textbf{98.23}\pm0.15$  & $\textbf{98.07}\pm0.25$ \\
    % \bottomrule
    \hline
  \end{tabular}}
  \end{center}
  \vspace{-0.3cm}
\end{table*}
The segmentation bottleneck within our P-CBM serves to dissect the input image into semantically meaningful regions, facilitating property concept prediction by the perceiver. Specifically, the perceiver's input comprises the concatenation of segmentation concepts provided by the observer and the original image. This is crucial because predefined property concepts, such as organ visibility, depend not only on the shape of the region of interest but also on its texture. For example, the property concept "clear stomach bubble" is described as "full and salient" with a "clear boundary"~\citep{wu2017fuiqa}. This motivates us to consider both texture and shape information in the perceiver.

Table~\ref{tab:fusion} presents ablation studies on information fusion within the perceiver. These experiments involved retraining the perceiver within the model. To isolate the impact of shape information alone, we replaced the $p(s_i | x)$ in Eq.~\ref{eq1} with the filled bounding box of the segmentation. Conversely, we examined the effect of solely feeding segmentation concepts to the perceiver without texture information. Additionally, we trained a perceiver using 'hard' segmentation concepts, i.e., replacing the $p(s_i | x)$ in Eq.~\ref{eq1} with the one-hot encoding of the predicted segmentation mask.

The results underscore the significance of fusing shape and texture information for passing to the perceiver. Furthermore, we observed that employing soft segmentation concepts enhanced model performance, indicating better calibration compared to binarized ones.

\subsubsection{Ablations on bottlenecks.}
We conducted ablation studies to evaluate the component contribution of P-CBM. The result in Table~\ref{tab:ablation} shows that all of our three technical contributions, i.e., the semantic bottleneck (SB) and the property bottleneck (CB) architecture contribute to the model performance.
% \subsubsection{Transfer learning in perceiver.}
% Fig.~\ref{fig:trainingcurve} illustrates the training curve of the perceiver net learning organ concepts, which proves that transfer learning helps the model converge at a faster speed. 
% \begin{figure}
%     \centering
%     \includegraphics[width=0.6\linewidth]{figures/training_curves.jpg}
%     \caption{Training curve of the perceiver with and without ImageNet initialization. }
%     \label{fig:trainingcurve}
% \end{figure}

\subsubsection{Model complexity.}
We evaluate the model complexity with three metrics: model parameters (Params), the floating point operations per second (FLOPs), and the inference time of the model for each image counted in milliseconds. The experiment was conducted on a Quadro RTX A6000 GPU with a batch size of 32. The Params and FLOPs of the model were evaluated by thop~\footnote{https://github.com/Lyken17/pytorch-OpCounter}.    

The full P-CBM took approximately 5 hours to be trained on a single Quadro RTX A6000 GPU for one split. The model parameter size is 66.32M, and FLOPs is 3954.70G when the batch size is 32. Although including a segmentation network, our model can still infer one image within 50 ms. This indicates a potential clinical application of our model in guiding the operators to capture standard planes in real-time since a typical refresh rate of a modern ultrasound scanner is 12-30 frames per second~\citep{fulgham2021physical} while our model can infer on 20 frames per second.  

\section{Conclusion}
\label{sec:conclusion}

In this paper, we propose progressive concept bottleneck model architecture, which consists of an observer \textit{\textbf{``seeing"}} segmentation concepts from input images, a perceiver \textit{\textbf{``conceiving"}} property concepts from the segmentation, and a predictor \textit{\textbf{``concluding"}} based on the property. Our segmentation concept is learned from supervised segmentation, which needs extra annotation. As future research, the segmentation bottleneck might be replaced by attention maps~\citep{fukui2019attention} learned from other models, saliency maps from eye trackers~\citep{rong2021human}, or by using methods from weak labeling segmentation~\citep{zhang2020survey}. Our annotations are relatively dense. 3rd-trimester ultrasound scans are challenging and give poor image quality. We aim to use the model not just to predict image quality, but to guide the clinician to acquire images of better quality, which would have a substantial clinical impact. This is also why we find the annotation burden for our use case worthwhile. We also note that recent foundation models~\citep{kirillov2023segment} may reduce the cost of segmentation annotation going forward. 

We demonstrate our model in the real-world scenario of decision support for assessing the quality of fetal ultrasound scans. This is challenging because of poor image quality due to image acquisition being difficult as well. Our experiments demonstrate that our model surpasses baseline models including previous CBMs and state-of-the-art classification models at recognizing ultrasound standard planes. Experiments show that our P-CBM provides correct and consistent explanations, which is useful to sonographers in screening guidance and downstream image analysis.

\section*{References}
\addcontentsline{toc}{section}{Reference}
\begingroup
\let\clearpage\relax
\vspace{-1.5cm}
\bibliographystyle{splncs04}
\bibliography{refs}
\endgroup

\end{document}